\documentclass[runningheads]{llncs}
\usepackage{graphicx}
\usepackage{cite}
\usepackage{amsmath,amssymb,amsfonts}
\usepackage{algorithmic}
\usepackage{graphicx}
\usepackage{textcomp}
\usepackage{adjustbox}
\usepackage{multirow}
\usepackage{bbold}
\usepackage{siunitx}
\usepackage{hyperref}       
\usepackage{colortbl}
\usepackage{xcolor}
\usepackage{sidecap}
\hypersetup{
    colorlinks,
    linkcolor={blue!50!black},
    citecolor={blue!50!black},
    urlcolor={blue!50!black}
}
\usepackage{amsmath, bm}
\usepackage{hyperref}
\usepackage{marvosym}

\usepackage{pifont}

\usepackage{floatrow}
\newfloatcommand{capbtabbox}{table}[][\FBwidth]

\usepackage{setspace}

\begin{document}

%

\title{A Review of 3D Reconstruction Techniques for Deformable Tissues in Robotic Surgery}
\titlerunning{A Review of 3D Reconstruction for Deformable Tissues in Robotic Surgery}

\author{Mengya Xu\inst{* 1,2} \and
Ziqi Guo\inst{* 1} \and
An Wang\inst{1} \and
Long Bai\inst{1,2} \and
Hongliang Ren\inst{\dag 1,2,3}}

\authorrunning{Xu et al.}

\institute{Dept. of Electronic Engineering, The Chinese University of Hong Kong (CUHK), Hong Kong SAR, China \and CUHK Shenzhen Research Institute, Shenzhen, China \and Dept. of Biomedical Engineering, National University of Singapore, Singapore  \\
\email{hlren@ee.cuhk.edu.hk}}

\maketitle              
\begin{abstract}
As a crucial and intricate task in robotic minimally invasive surgery, reconstructing surgical scenes using stereo or monocular endoscopic video holds immense potential for clinical applications. NeRF-based techniques have recently garnered attention for the ability to reconstruct scenes implicitly. On the other hand, Gaussian splatting-based 3D-GS represents scenes explicitly using 3D Gaussians and projects them onto a 2D plane as a replacement for the complex volume rendering in NeRF. However, these methods face challenges regarding surgical scene reconstruction, such as slow inference, dynamic scenes, and surgical tool occlusion. This work explores and reviews state-of-the-art (SOTA) approaches, discussing their innovations and implementation principles. Furthermore, we replicate the models and conduct testing and evaluation on two datasets. The test results demonstrate that with advancements in these techniques, achieving real-time, high-quality reconstructions becomes feasible. The code is available at: \href{https://github.com/Epsilon404/surgicalnerf}{https://github.com/Epsilon404/ surgicalnerf}.

\end{abstract}

\renewcommand{\thefootnote}{}
\footnotetext{\inst{*} Authors contributed equally to this work.}
\footnotetext{\inst{\dag} Corresponding author.}

\section{Introduction}
Reconstruction of surgical scenes from stereo or monocular endoscopic video is a significant and complicated mission in robotic minimally invasive surgery, which could implement clinical applications such as augmented reality in surgical environments and precise surgical navigation \cite{knappe2003position,qian2019review}. However, the most challenging tasks not only lie in a large amount of consumption of computation time and resources but also exist in the endoscopic view with limited viewing directions and the dynamic scene with non-rigid deforming tissues, noticeable lighting variations, and surgical instruments occlusions \cite{wang2022endonerf,batlle2023lightneus,psychogyios2023REIM}. 

After the emergence of NeRF \cite{mildenhall2020nerf}, there has been an increasing number of studies focusing on implicit reconstruction techniques inspired by NeRF to enhance its functionality and broaden its range of applications. Among them, EndoNeRF \cite{wang2022endonerf} sets the first precedent for applying it to robotic surgery by incorporating the neural radiance field for deformable tissue reconstruction to solve the above challenges. It utilizes an innovative ray sampling method that enhances the probability of casting the ray on pixels with high occluded frequency, aiming at avoiding the effects of occluding tools and removing them. Then, reconstruct the scene by integrating D-NeRF \cite{pumarola2021d}. 

Since the first appearance of deformable tissue reconstruction, the number of works exceeding EndoNeRF significantly increased. EndoSurf \cite{zha2023endosurf} and Neural LerPlane \cite{yang2023lerplane} are two notable studies among them to reconstruct a smooth surface and significantly reduce the reconstruction time, respectively. From the result of NeuS \cite{wang2021neus}, a signed distance function (SDF) works well in restoring a smoother surface than a density field in NeRF \cite{mildenhall2020nerf}. As a result, EndoSurf \cite{zha2023endosurf} was initiated to reconstruct smooth surfaces in a deformation surgical scene using the SDF. It employs three networks to accomplish the task: a deformation network converting points from observation space into canonical space, an SDF network precisely depicting the geometry of tissue surface, and a radiance network learning the color attributes of surface points. By incorporating these enhanced network structures, EndoSurf significantly improves reconstruction for deformable tissues. 
Meanwhile, enhancing training speed for real-time reconstruction during surgical procedures is crucial. To address this challenge, LerPlane \cite{yang2023lerplane} adopts a novel approach by dividing the 4D scene into two components: a static field that remains constant over time and a dynamic field that captures temporal changes. Each component is divided into multiple 2D planes: three planes represent spatial points in the static field. In comparison, another set of three planes captures temporal variations of spatial points within the dynamic field. This decomposition simplifies the projection process for each spatial point onto six 2D planes and facilitates the integration of features. Consequently, it reduces the complexity associated with deformation reconstruction, significantly reducing training time. This advancement offers promising prospects for achieving real-time reconstruction in robotic surgery.

Recently, 3D Gaussian Splatting \cite{kerbl3Dgaussians} has taken a different route from NeRF in scene reconstruction. It represents the scene as explicit 3D Gaussians and directly projects them onto the 2D plane, known as differentiable splatting \cite{Yifan_2019}, to replace the complex volume rendering in NeRF. Therefore, 3D-GS has an observable 3D scene and a real-time rendering speed. Based on it, 4D Gaussians Splatting \cite{wu20234dgaussians} imports time information to expand it to dynamic scenes. 4D-GS introduces a deformation field to predict the motion and variation of each 3D Gaussian at a specific time and splats 3D Gaussians to render the image.

In this work, we review 4 methods of surgical scene reconstruction, including EndoNeRF \cite{wang2022endonerf}, EndoSurf \cite{zha2023endosurf}, LerPlane \cite{yang2023lerplane} and 4D-GS \cite{wu20234dgaussians}, then reproduce their models and results, observe and evaluate their performance on not only the basic EndoNeRF dataset \cite{wang2022endonerf}, also the additional StereoMIS dataset \cite{hayoz2023stereomis} and C3VD dataset \cite{bobrow2023}. Our contributions are:
\begin{itemize}
\item We evaluate the SOTA methods EndoNeRF, EndoSurf, LerPlane, and 4D-GS in terms of training time, GPU usage, and performance on three datasets.
\item We compare the NeRF-based methods with Gaussian Splatting, which no work before us has investigated.
\item We discover the domain gap between natural and surgical environments leads to a reduced generalization performance of 4D-GS when applying it to surgical scenes, which underscores the need for innovative approaches to address the challenge effectively.

\end{itemize}

\section{Methodology}

In this section, we first review the principles of two basic models, NeRF and 3D Gaussian Splatting in Sec. \ref{preliminaries}, and then introduce the implementation of the four methods in Sec. \ref{sec:endonerf} -- Sec. \ref{sec:4dgs}.

\subsection{Preliminaries}\label{preliminaries}

\subsubsection{NeRF: Neural Radiance Field}
NeRF \cite{mildenhall2020nerf} utilizes a function $F_\Theta$ to map each spacial point location $\textbf{x}=(x,y,z)$ and viewing direction $\textbf{d}=(\theta,\phi)$ to the output point color and volume density $(\textbf{c},\sigma)$, i.e., $F_\Theta = (x,y,z,\theta,\phi) \mapsto (\textbf{c},\sigma)$. 
Then it defines a camera ray by $\textbf{r}(t) = \textbf{o}+t\textbf{d}$, where the ray is emitted from $\textbf{o}$ in the direction of $\textbf{d}$ and reaches $\textbf{r}(t)$.
Finally uses the classical volume rendering \cite{volumerendering} to predict the pixel color $\hat{C}(\textbf{r})$ and depth $\hat{D}(\textbf{r})$: $\hat{C}(\textbf{r})=\int_{t_n}^{t_f}T(t)\sigma(\textbf{r}(t))\textbf{c}(\textbf{r}(t))dt, 
    \hat{D}(\textbf{r})=\int_{t_n}^{t_f}T(t)\sigma(\textbf{r}(t))tdt, 
    T(t)=\exp(-\int_{t_n}^{t}\sigma(\textbf{r}(s))ds)$.

\subsubsection{3D Gaussian Splatting}
Unlike NeRF, 3D-GS \cite{kerbl3Dgaussians} represents a scene by explicit 3D Gaussian ellipsoids. Each 3D Gaussian has four attributes to be optimized: center point $\mathcal X$, covariance matrix $\Sigma$, opacity $\alpha$, and color $c$. 
A 3D Gaussian can then be represented by: $G(X) = \exp(-\frac{1}{2} \mathcal X^T\Sigma^{-1}\mathcal X)$.
To render the image on novel views, it first computes the 2D covariance matrix $\Sigma' = JW\Sigma W^TJ^T$ to be an attribute of the projected ellipse on the 2D plane, where $J$ is the Jacobian matrix of projective transformation, and $W$ is the viewing transformation. Then, a pixel color can be calculated with: $\hat{C} = \sum_i c_i \alpha_i \prod_{j=1}^{i-1}(1-\alpha_j)$.

\subsection{EndoNeRF: Endoscopic NeRF Reconstruction}\label{sec:endonerf}
To represent a deformable tissue scene, EndoNeRF \cite{wang2022endonerf} firstly uses the modeling process in D-NeRF \cite{pumarola2021d} to build two fields: a static canonical field and a time-dependent deformation field. The canonical field follows the same function $F_\Theta$ as NeRF~\cite{mildenhall2020nerf} in Sec.~\ref{preliminaries}, while the deformation field $G_\Phi$ maps the location \textbf{x} in the canonical field and time $t$ to the distance from the static \textbf{x} to the point \textbf{x} at time $t$. Thus, the color and density of one point \textbf{x} at a certain time $t$ can be gained by $(\textbf{c},\sigma)=F_\Theta(\textbf{x}+G_\Phi(\textbf{x},t),\textbf{d})$. 

Next is to sample rays on a randomly chosen frame. Following the uniform random sampling strategy in NeRF is not conducive to removing tool occlusion. Therefore, EndoNeRF constructs importance maps $\mathcal V_i$ where $i$ is the frame index to guide the casting of rays. 

\begin{equation}\label{eq:endonerf1}
    \mathcal V_i=\Lambda \otimes (\textbf{1}-M_i), \\
    \Lambda=\left(\textbf{1}+\frac{\sum_j M_j}{\|\sum_j M_j\|_F} \right), \\
    \hat{\mathcal{V}_i}=\frac{\mathcal{V}_i}{\|\mathcal{V}_i\|_F}
\end{equation}

In Eq. \ref{eq:endonerf1}, $M_i$ is the tool mask of frame $i$ where tool pixels are marked as 1, the constant $\Lambda$ gives a higher importance on the tool occluded pixels, and $\otimes$ is element-wise multiplication. Then, by normalization, $\hat{\mathcal{V}_i}$ gives the probability mass function to sample rays where the probability of casting rays on tool pixels at this time frame $i$ is zero. 

The sampling point step leverages the estimated stereo depth to generate a Gaussian distribution sampling strategy, which samples more points near the surface of the tissue: $\delta(s;u,v,i) = \exp(-{(s-\textbf{D}_i(u,v))^2}/{2\xi^2})$. Here, $s$ is the distance on the ray $\textbf{r}(s)$ and $\textbf{D}_i$ is the depth of pixel $(u,v)$. Then, by classical volume rendering \cite{volumerendering} as in NeRF, it predicts the color and depth to compute the loss function. 

\subsection{EndoSurf: Endoscope-based Surface Reconstruction}\label{sec:endosurf}
The novelty of EndoSurf \cite{zha2023endosurf} is reconstructing the tissue surface and texture. It defines three neural fields: a deformation field $\mathbf{\Psi}_d$ for the deformable scene, an SDF field $\mathbf{\Psi}_s$ for the surface, and a radiance field $\mathbf{\Psi}_r$ for surface texture, to solve the problem. Similar with EndoNeRF \cite{wang2022endonerf}, the deformation field maps a point $\textbf{x}_o$ at time $t$ to the displacement between $\textbf{x}_o$ and its corresponding point in canonical space $\textbf{x}_c$, i.e., $\Delta\textbf{x} = \mathbf{\Psi}_d (\textbf{x}_o,t)$, and $\textbf{x}_c=\textbf{x}_o+\Delta\textbf{x}$. Inspired by NeuS \cite{wang2021neus}, the SDF field takes canonical point $\textbf{x}_c$ as input and takes the signed distance function $\rho$ with a geometry feature vector \textbf{f} of the point as outputs, i.e., $(\rho,\textbf{f})=\mathbf{\Psi}_s(\textbf{x}_c)$. Here $\rho$ has to be positive when $\textbf{x}_c$ is between the camera and the surface and otherwise negative. In this setting, the surface needed to be reconstructed is represented by $\mathcal{S}=\{\textbf{x}|\mathbf{\Psi}_s(\textbf{x})=0\}$, and one can calculate the surface normal $\textbf{n}_c$ of a surface point $\textbf{p}_c$ by the gradient: $\textbf{n}_c = \nabla\mathbf{\Psi}_s(\textbf{p}_c)$. For the radiance field, it outputs the pixel color $\textbf{c}_c$ with input $(\textbf{x}_c,\textbf{v}_c,\textbf{n}_c,\textbf{f})$, where the parameters are spacial coordinates, the viewing direction, the surface normal, and the geometry feature vector.

With the predicted signed density function $\rho_i$ and color $\textbf{c}_i$ of sampled points $\textbf{x}_i$ on ray $\textbf{r}(h)$ at time $t$, it is able to conduct unbiased volume rendering \cite{wang2021neus} to estimate the ray color and depth:
\begin{equation}
    \hat{\textbf{C}}(\textbf{r}(h))=\sum_i \left(\prod_{j=1}^{i-1}(1-\alpha_j)\right)\alpha_i\textbf{c}_i,\\
    \hat{\textbf{D}}(\textbf{r}(h))=\sum_i \left(\prod_{j=1}^{i-1}(1-\alpha_j)\right)\alpha_i h_i
\end{equation}
where $\alpha_i=\max\left(1-\frac{1+\exp(-\rho_i/s)}{1+\exp(-\rho_{i+1}/s)}, 0 \right)$ is the opacity of each point on the ray. 

Ultimately, it sets the loss functions to optimize the rendered images and the SDF. The rendering loss is defined by $\lambda_1\mathcal{L}_{c}+\lambda_2\mathcal{L}_{d}$.
\begin{equation}
    \mathcal{L}_{c}=\sum_{\textbf{r}}\|M(\textbf{r})(\hat{\textbf{C}}(\textbf{r})-\textbf{C}(\textbf{r}))\|_1,\\
    \mathcal{L}_{d}=\sum_{\textbf{r}}\|M(\textbf{r})(\hat{\textbf{D}}(\textbf{r})-\textbf{D}(\textbf{r}))\|_1
\end{equation}
Here $M$ represents the tool mask and \textbf{C}, \textbf{D} are ground truth color and depth. Then it optimizes the SDF field by four loss functions: the Eikonal loss \cite{gropp2020implicit}, the SDF loss that requires the SDF outputs of surface points to be zero, the visible loss that generates a correct surface direction, and the smoothness loss that gives a smooth surface.

\subsection{LerPlane: Linear Interpolation Plane}\label{sec:lerplane}
The reason why LerPlane \cite{yang2023lerplane} can rapidly reconstruct a surgical scene is the idea of decomposing the 4D scene into six explicit 2D planes similar to \cite{kplanes_2023}, which reduces the complexity from $O(N^4)$ to $O(N^2)$, and shrinks the neural network to a tiny MLP to accelerate the training. Specifically, it constructs two fields with three planes each to represent a deformable surgical scene. Three space planes $XY, YZ, XZ$ form the static field, and three time-dependent planes $XT, YT, ZT$ constitute the dynamic field. The total $6$ planes are orthogonal to each other, resulting in a simple projection for a 4D point onto each plane. 

To remove tool occlusion, a spatiotemporal importance sampling strategy is utilized for ray casting. With the tool mask $\textbf{\textit{M}}_i$ and input image $\textbf{\textit{I}}_i$ of the $i^{th}$ frame, a weight map $\textbf{W}_i$ similar to the importance map of EndoNeRF \cite{wang2022endonerf} is generated by:
\begin{equation}
    \textbf{W}_i=\min[\frac{1}{3}\max_{j\in(i-n,i+n)}(\|\textbf{\textit{I}}_i\otimes\textbf{\textit{M}}_i-\textbf{\textit{I}}_j\otimes\textbf{\textit{M}}_j\|_1),\alpha]\otimes\mathbf{\Omega}_i,
    \mathbf{\Omega}_i=\beta\left(\frac{\textbf{\textit{M}}_i T}{\sum_{i=1}^T \textbf{\textit{M}}_i}\right)
\end{equation}
where $\otimes$ is element-wise multiplication and hyperparameters $\alpha$ and $\beta$ represent a lower bound and a balancing parameter respectively. Then, on the frequently occluded pixels, the $\mathbf{\Omega}$ value will be higher, leading to a higher probability of sampling rays on these pixels.

With the sampled ray and 4D point, the 2D features are extracted by projecting the point on six 2D planes and utilizing the bilinear interpolation method. Then fuse the six features into the final feature vector fed into the MLP, which estimates the color and density $(\textbf{c},\sigma)$, and renders the color and depth by volume rendering~\cite{volumerendering}. 
Optimizing the MLP and fields leverages not only the color and depth loss but also the total variation loss and smooth time loss that ensure the similarity of adjacent frames.

\subsection{4D-GS: 4D Gaussian Splatting}\label{sec:4dgs}
Similar to the above methods of modeling deformable scenes, 4D-GS \cite{wu20234dgaussians} is mainly aiming at optimizing a deformation field $\mathcal{F}$ to output the new states of 3D Gaussians in the space at a specific time $t$. Such a deformation field is separated into two parts in the implementation: multi-resolution neural voxels that extract the features on voxel planes and a tiny MLP that outputs the information of deformed 3D Gaussians by decoding the features.

Based on the fact that the 3D Gaussians with proximal space positions have similar states and that one 3D Gaussian will have akin features in adjacent timestamps, it utilizes a HexPlane module with multi-resolution to encode the information, including time $t$ of all 3D Gaussians. Like the idea in LerPlane \cite{yang2023lerplane}, the HexPlane module uses six 2D voxel planes with interpolation to extract and fuse features. 
\begin{equation}
    f=\bigcup\prod interp(R(i,j)), (i,j)\in\{(x,y),(y,z),(x,z),(x,t),(y,t),(z,t)\}
\end{equation}

With the features in voxels, the required parameters, including the variation of location, rotation, scaling, opacity, and color, i.e., $(\Delta\mathcal{X}, \Delta r, \Delta s, \sigma, \mathcal{C})$, can be decoded by a tiny MLP. A new state of each 3D Gaussians is then represented by $\mathcal{S}'=\mathcal{F}(\mathcal{S},t)=(\mathcal{X}+\Delta\mathcal{X}, r+\Delta r, s+\Delta s, \sigma, \mathcal{C})$, and the differential splatting \cite{Yifan_2019} is exploited to render the final color $\hat{C}=\mathcal{G}(\mathcal{S}'|R,T)$ with the view-matrix $[R,T]$. Finally, it uses color reconstruction loss and total variation loss to optimize. 

\section{Experiments}
\subsection{Dataset Description} 
We evaluate the models on 3 public datasets, \textbf{EndoNeRF dataset}~\cite{wang2022endonerf}, \textbf{StereoMIS dataset}~\cite{hayoz2023stereomis}, and \textbf{C3VD} \cite{bobrow2023}.

\textbf{EndoNeRF dataset} gives two endoscopic scenes generated from in-house DaVinci robotic surgery scenes. The video of each scene is captured by a single-viewpoint stereo camera. Each image has a resolution of $640\times512$, with a corresponding depth map and a binary mask of the surgical tool. The depth map is estimated by \cite{Li_2021_sttr}, and the tool mask is manually labeled in left camera images.

\textbf{StereoMIS dataset} is captured from the da Vinci Xi robotic surgery scenes. Each of the 11 scenes includes a stereo video from a single viewpoint and a set of binary tool masks. The video data is processed into left and right camera view image sets, with each image resolution $640\times512$.

\textbf{C3VD} is a colonoscopy 3D video dataset with totally 22 video sequences. The images have a resolution of $640\times512$ and the depth map is generated from optical images by a Generative Adversarial Network (GAN).

\subsection{Implementation details}
We train and evaluate the models on the same platform with Ubuntu 20.04 and one RTX3090 GPU. We split the training data into train and test sets with the quantity ratio 7:1. Specifically, in the data sequence, after grouping $8$ images, the first $7$ images are added to the training set, and the last one is added to the testing set. We utilize the Depth Anything small-sized model \cite{yang2024depthanything} to generate a set of coarse depth maps for the StereoMIS dataset. We train each model until their respective convergence.
For the EndoNeRF model, we train it in $100K$ iterations for about 5 hours. 
For the EndoSurf model, we train it in $100K$ iterations for about 10 hours. 
For the LerPlane model, we train it in $32K$ iterations for 12 minutes. 
For the 4D-GS model, we train it in $6K$ iterations for 5 minutes. 
The rest of the experimental settings retain the default values for each model. Finally, we use the image quality evaluation index to appraise the reconstruction performance of each model, including PSNR, SSIM, and LPIPS. These metrics present the similarity between synthesized and test set images, giving the quantitative results of the models on two datasets.

\section{Results and Evaluation}
\begin{figure}[htb]
\centering
\caption{Visualization Results of 4 models on $2$ datasets. EndoNeRF, EndoSurf, and LerPlane on the first dataset give good 3D results. Still, LerPlane on the StereoMIS dataset cannot restore the original color, and 4D-GS presents poor performance on both endoscopic datasets.}
\includegraphics[width=1\linewidth]{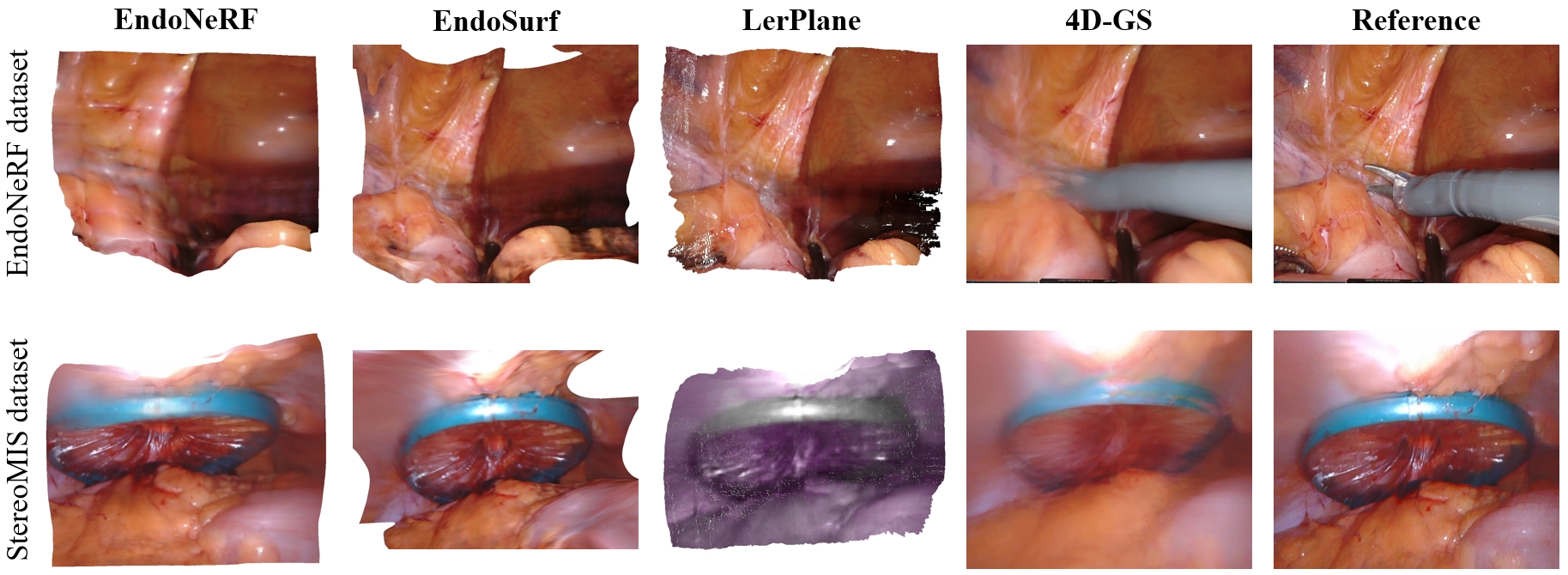}
\label{fig1}
\end{figure}

We evaluate and compare the 4 models EndoNeRF \cite{wang2022endonerf}, EndoSurf \cite{zha2023endosurf}, LerPlane \cite{yang2023lerplane} and 4D-GS \cite{wu20234dgaussians} on 3 datasets EndoNeRF \cite{wang2022endonerf}, StereoMIS \cite{hayoz2023stereomis} and C3VD \cite{bobrow2023} using metrics of PNSR, SSIM and LPIPS, together with training time, inference time and GPU usage. We train each model until respective convergence. The evaluation and comparison are shown in Table~\ref{table1} and Table~\ref{table2}. Fig.~\ref{fig1} shows the visualization results.

\begin{table}[htb]
\centering
\caption{Quantitative Results of $4$ models on $3$ datasets. The ones in bold are the best value, and the underlined ones take second place. EndoNeRF and EndoSurf give better reconstruction results.}

\scalebox{0.89}{
\begin{tabular}{c|ccc|ccc|ccc}
\hline
\multirow{2}{*}{Models} & \multicolumn{3}{c|}{EndoNeRF dataset\cite{wang2022endonerf}} & \multicolumn{3}{c|}{StereoMIS dataset\cite{hayoz2023stereomis}} & \multicolumn{3}{c}{C3VD dataset\cite{bobrow2023}} \\ \cline{2-10}
                                & PSNR↑      & SSIM↑      & LPIPS↓      & PSNR↑       & SSIM↑      & LPIPS↓     & PSNR↑       & SSIM↑      & LPIPS↓ \\ \hline
EndoNeRF\cite{wang2022endonerf} & 27.077     &  0.900     &  \underline{0.107}      & \textbf{31.511}      &   \textbf{0.832}     &   \textbf{0.190}      & \textbf{36.759}       &  \textbf{0.886}      &     \textbf{0.214}      \\
EndoSurf\cite{zha2023endosurf}  &  \textbf{34.795}     &   \textbf{0.945}     &  0.119      & \underline{28.417}      &  \underline{0.818}     &  \underline{0.368}      &  \underline{33.192}      &  \underline{0.868}      &     \underline{0.346}     \\
LerPlane\cite{yang2023lerplane} & \underline{34.643}     &  \underline{0.922}     &   \textbf{0.072}      & 17.526      &  0.741     &  0.379      &   16.914     &  0.845      &      0.348 \\
4D-GS\cite{wu20234dgaussians}   & 22.832     &  0.827     &  0.368      & 19.202      &  0.756     &  0.472      &  21.352      & 0.865       &     0.437    \\ \hline
\end{tabular}}
\label{table1}
\end{table}
\begin{table}[htb]
\centering
\caption{Results for proof of real-time performance evaluation. 4D-GS consumes the shortest time and is closest to real-time rendering.}
\setlength{\tabcolsep}{12pt}
\begin{tabular}{c|ccc}
\hline
Models   & Training Time & Inference Time & GPU Usage \\ \hline
EndoNeRF\cite{wang2022endonerf} &  6 h &  8585.3 ms   &  \underline{8 GB}  \\
EndoSurf\cite{zha2023endosurf} &   10 h &  33476.6 ms   &  19 GB        \\
LerPlane\cite{yang2023lerplane} &   \underline{12 min} &  \underline{601.3 ms}    &  22 GB        \\
4D-GS\cite{wu20234dgaussians}    &   \textbf{5 min} &  \textbf{18.3 ms}   &   \textbf{4 GB}      \\ \hline
\end{tabular}
\label{table2}
\end{table}

From the experiment results, we can observe that (1) Training time: EndoSurf $\textgreater$ EndoNeRF $\gg$ LerPlane $\textgreater$ 4D-GS. The reason EndoNeRF and EndoSurf require more time is that they incorporate time information, which increases complexity. In contrast, the other two models employ a 2D plane representation, effectively reducing complexity and decreasing inference and training times. The inference time has the same trend as the training time, meaning that EndoNeRF and EndoSurf are difficult to use for real-time rendering, but LerPlane and 4D-GS have the potential for it. (2) GPU usage: LerPlane $\textgreater$ EndoSurf $\textgreater$ EndoNeRF $\textgreater$ 4D-GS. EndoSurf consumes more GPU memory as it requires an additional field to locate the tissue surface. LerPlane's higher GPU usage may be attributed to the experiment's excessive multi-resolution setting and the subsequent large number of spatial points it generates. Reducing the number of explicit Gaussians and instead utilizing a smaller set of implicit points could potentially minimize GPU usage for 4D-GS. (3) Performances: The performance of LerPlane on the EndoNeRF dataset is comparable to that of EndoSurf, although it falls slightly short due to the inherent variability within the margin of error. LerPlane's ability to attain these results in a shorter time underscores the efficacy of its acceleration strategy. The disparate trends observed between the two datasets may be attributed to the distinct characteristics of the input data. Specifically, the EndoNeRF dataset presents a more deformable scene, whereas the StereoMIS and C3VD datasets exhibit less deformation. The subpar performance of LerPlane on the latter dataset likely arises from its inability to capture features within a more static scene context effectively.

\section{Conclusion}
In conclusion, this work summarizes and reviews the existing state-of-the-art models in surgical scene reconstruction. EndoNeRF introduces NeRF for the first endoscopic reconstruction, while EndoSurf restores a smooth surface, LerPlane dramatically increases training speed, and 4D-GS takes advantage of Gaussians to reconstruct the scene explicitly. 
4D-GS performs worse on the surgical scene than the natural scene, implying the domain gap for transferring it directly to surgical scenes: (a) Due to the limited viewing direction, it cannot restore the initial Gaussians, (b) Surgical scenes with large deformations result in poor generalization, (c) Removing surgical tools also requires a sampling strategy similar to the previous models. We believe future studies will enable 4D-GS to achieve successful surgical scene reconstruction, closer to the purpose of real-time rendering. Meanwhile, a series of recent methodologies on endoscopic reconstruction related to GS and foundation models can be found at~\cite{cui2024endodac,cui2024surgical,huang2024endo,huang2024registering,li2024endosparse,liu2024endogaussian,liu2024lgs,yang2024deform3dgs,yang2024efficient,zhao2024hfgs,zhu2024endogs}.

\begin{credits}
\subsubsection{\ackname} This work was supported by Hong Kong RGC CRF C4026-21G, GRF 14211420 \& 14203323, Shenzhen-Hong Kong-Macau Technology Research Programme (Type C) STIC Grant SGDX20210823103535014 (202108233000303) and the Key Project 2021B1515120035 (B.02.21.00101) of the Regional Joint Fund Project of the Basic and Applied Research Fund of Guangdong Province.
\subsubsection{\discintname}
The authors have no competing interests to declare.
\end{credits}

\bibliography{mybib}{}
\bibliographystyle{splncs04}
\end{document}